\documentclass[10pt]{article}

\usepackage{iclr2027_conference,times}
\usepackage{booktabs}
\usepackage{amsmath,amssymb}
\usepackage{graphicx}
\usepackage{hyperref}
\usepackage{xcolor}
\usepackage{microtype}
\usepackage{enumitem}

\newcommand{\bench}{DynGraphAgentBench}
\newcommand{\capmetric}[1]{\mathrm{Capture}@#1}

\iclrfinalcopy \title{\bench: A Benchmark for Agentic Lifecycle Control in Dynamic Graph Anomaly Detection}
\author{\mdseries Yuwei Han, Lingwei Wei, Wooseong Yang, Liangjie Huang \\ \mdseries Liancheng Fang, Huanhuan Ma, Philip S. Yu}
\begin{document}
\maketitle

\begin{abstract}
Dynamic graph anomaly detection requires repeated decisions as graph structure and class prevalence drift, yet detector benchmarks usually score a fixed pipeline after current labels are known. We introduce DynGraphAgentBench, an executable benchmark for agentic lifecycle control under delayed feedback. It comprises seven temporal graph datasets with node- and edge-level anomaly tasks, eleven selectable detectors, and eight chronological deployment windows per dataset. In each window, a controller sees only time-causal aggregate context, registered model cards, and its own matured history. It must choose a detector before current-window training or candidate scores exist. A sandboxed executor trains the chosen architecture on mature data, scores a hidden deployment window, and releases the outcome after a one-window delay. A deterministic verifier checks decision timing, leakage guards, legal actions, training scope, and persisted artifacts. We measure detection utility with average precision and capture at fixed review depth, and characterize adaptation through model switches and compute. Complete eight-window trajectories from two primary controllers and a no-memory reference on four datasets, together with three additional controllers on three datasets, expose useful, costly, and ineffective reactions to delayed evidence without granting an exhaustive current-window oracle.
\end{abstract}

\section{Introduction}

Interaction graphs evolve continuously after deployment. New entities enter the graph, interaction patterns change, and anomaly prevalence can shift substantially over time. Meanwhile, reliable labels often arrive only after a delay, as investigations and downstream outcomes are resolved after the original decision. A practical controller therefore needs to select and adapt detectors using only the information available at each point in time, rather than relying on future labels to identify the best-performing candidate.

This distinction is routinely blurred. Dynamic graph work has produced strong temporal encoders such as TGN and DyGFormer \citep{rossi2020tgn,yu2023dygformer}, while BAG provides a broad, unified anomaly-detection implementation \citep{hua2026bag}. Yet a benchmark that trains every candidate on the current split and then asks an agent to pick from current validation scores measures ranking, not model-selection or adaptation. It also makes an agent appear computationally efficient while an evaluator has already paid for exhaustive training.

\begin{figure*}[t]
\centering
\includegraphics[width=0.96\textwidth]{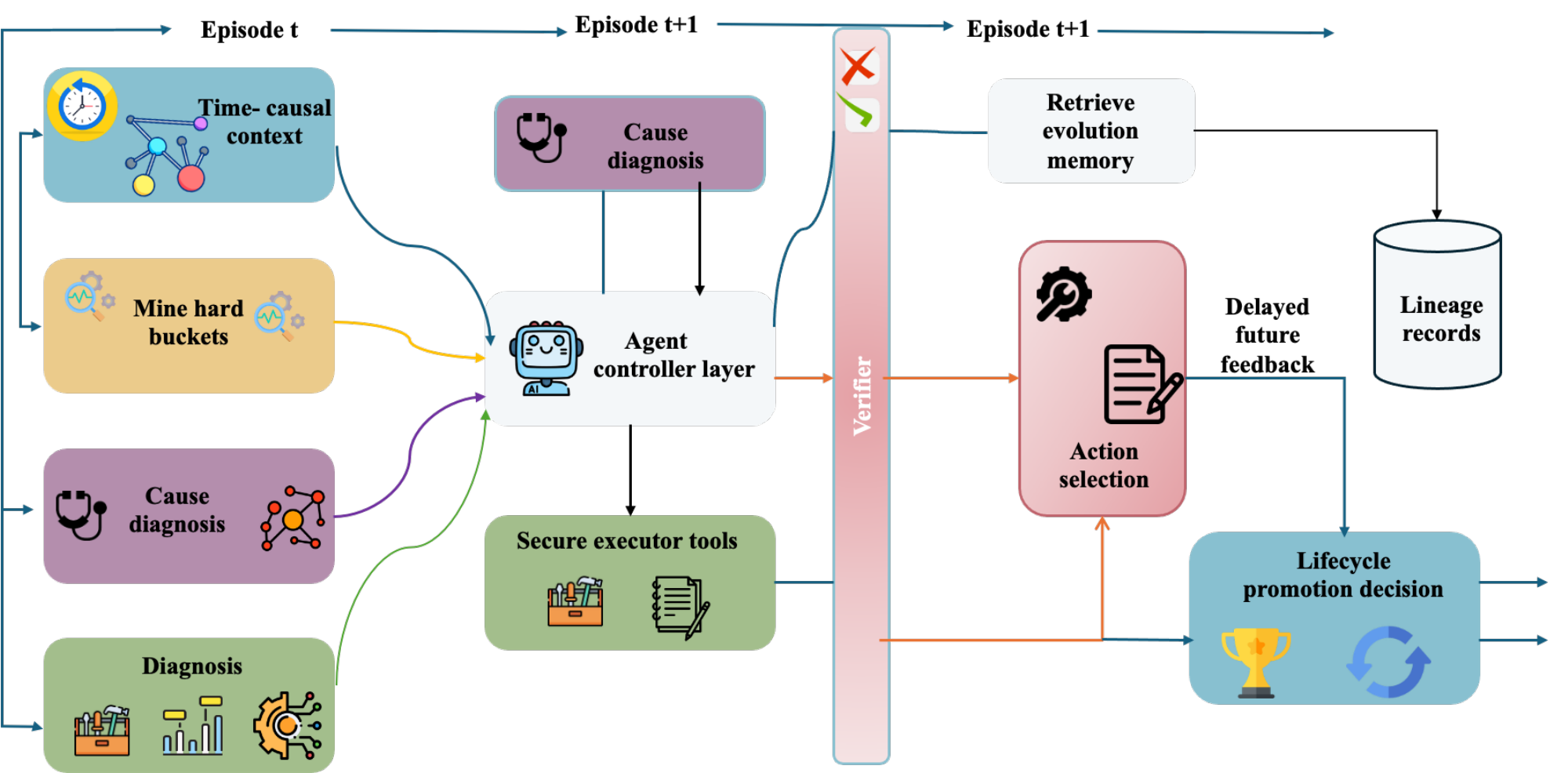}
\caption{Agent control loop overview. Time-aware context, hard-bucket mining, and diagnosis inform the controller. A verifier checks proposed actions before secure execution; delayed feedback and lineage records inform later episodes.}
\label{fig:agent-control-loop}
\end{figure*}

We instead impose an \emph{agent-first} order. In every deployment window, a controller receives dataset statistics, eleven model cards, its previous choice, and only outcomes whose labels have matured. It must select an architecture before any current-window training begins. The executor durably records the decision, trains the selected architecture from all mature history, and evaluates once on a hidden chronological window. A deterministic verifier checks the resulting artifacts and protocol state; the controller's rationale is never accepted as evidence of success. GPT-6 Astra and Claude Opus 5 controllers have separate histories and start afresh on each dataset. Figure~\ref{fig:agent-control-loop} gives an overview of the control loop and its delayed feedback.

Our benchmark validation asks three focused questions. \textbf{RQ1:} Can the protocol distinguish two agent-first policies on hidden windows? \textbf{RQ2:} Does it expose whether agents respond usefully when delayed evidence indicates that a previous choice degraded? \textbf{RQ3:} Does removing cross-window outcome memory measurably change selection, switching, or cost? These questions require both predictive and behavioral measurements: average precision (AP) and capture at fixed review depth quantify deployment utility; switch rate, response to negative feedback, legal-choice rate, and triggered GPU time characterize the controller.

Benchmark validity does not require either hosted controller to outperform the no-memory reference. A controller that persists with a weak choice, reacts ineffectively to delayed evidence, or incurs extra training cost is an informative outcome if the protocol exposes it reproducibly. Our validation criterion is therefore whether the suite produces auditable, behaviorally discriminating trajectories under a fixed information contract, not whether a particular proprietary agent establishes a new detector state of the art.

We make three contributions:
\begin{enumerate}[leftmargin=*]
    \item a leakage-safe, eight-window protocol that fixes the exact information available before selection and delays deployment outcomes by one complete window;
    \item a unified execution layer exposing a finite bank of nine temporal graph encoders and two graph--attribute fusion models through typed adapters on node- and edge-level tasks; and
    \item an artifact-backed evaluation of two hosted language agents and a no-memory reference on four datasets, plus three additional controllers on three datasets, without using an exhaustive current-window oracle.
\end{enumerate}

\paragraph{Benchmark design goals.}
\bench{} is designed around four criteria. \emph{Construct validity:} selection occurs before current candidate training, so the task measures a decision under uncertainty rather than retrospective ranking. \emph{Coverage:} seven defined temporal graph environments span node- and edge-level anomaly units, with eleven selectable architectures and two primary controllers. \emph{Auditability:} immutable manifests, pre-training decisions, and final-state checks expose temporal leakage and compute. \emph{Extensibility:} a typed dataset contract and candidate registry allow new streams, agents, and models without changing the evaluator.

\section{Agent-First Rolling Selection}

\subsection{Sequential decision problem}

Let $D_{\leq t}$ be all records observed by decision time $t$ and let $Y_{\leq t-d}$ denote labels that have matured after delay $d$. Each dataset is divided into an initial chronological bootstrap period followed by eight disjoint deployment windows $W_0,\ldots,W_7$. We fix $d$ to one complete window. At episode $e_k$, the newest deployment outcome available to the agent is therefore $W_{k-2}$; $W_{k-1}$ is still unresolved.

The agent observes
\begin{equation}
o_k=(s_k,\mathcal{M},a_{k-1},H_{\leq k-2}),
\end{equation}
where $s_k$ contains only aggregate mature train/validation counts and the number of unlabeled current examples, $\mathcal{M}$ is the fixed model bank, $a_{k-1}$ is the agent's preceding choice, and $H_{\leq k-2}$ contains its matured validation and deployment outcomes. It chooses one candidate $a_k\in\mathcal{M}$. The executor then trains $a_k$ using only mature labels and evaluates scores on $W_k$. No current label, current candidate metric, evaluator file, or competing agent state is present in $o_k$. The benchmark evaluates the state produced by the controller and executor, not a self-reported completion claim.

\subsection{Action and sandbox contract}

Each response must be a JSON object with a legal \texttt{candidate\_id}, an action in \{\texttt{initialize}, \texttt{keep}, \texttt{switch}\}, a rationale, and cited evidence from the payload. The executor normalizes the action label to the actual transition while leaving the originally reported action in the trace. A fail-closed payload checker rejects any key containing \texttt{test} or \texttt{oracle}. Agents have no shell, filesystem, network tools, evaluator mount, or access to one another. Training occurs behind the executor rather than inside the language-model sandbox.

This is intentionally narrower than autonomous machine-learning engineering benchmarks \citep{huang2024mlagentbench,chan2024mlebench}. The agent cannot write model code or tune arbitrary hyperparameters. Constraining the action to model selection makes causal ordering testable and makes two agents directly comparable.

We call this the \emph{controlled controller track}. It isolates sequential model choice under a fixed registry, tool surface, and budget. An open-ended \emph{native lifecycle track}, in which agents edit code, construct graphs, and manage long-running workflows, is complementary future work rather than a capability claimed by the present benchmark.

\paragraph{Final-state verification.}
A trajectory is protocol-valid only if every required check passes: the decision predates training, the candidate belongs to the registered bank, the payload and data slices pass leakage guards, training matches the declared selected-union-plus-reference scope, and the hidden deployment window is evaluated once. We report this strict conjunction separately from continuous detection utility (AP, fixed-depth capture, and AUROC). This separation follows the executable-workflow principle of \citet{chen2026chibench}, while using deterministic verification wherever the fraud setting permits it.

\paragraph{Two kinds of memory.}
Agent memory $M_k^{\mathrm{agent}}$ stores matured outcomes, prior selections, and experiment history across rolling windows. Detector memory $M_k^{\mathrm{detector}}$ is model state created during one training episode and discarded before the next; TGN's internal temporal memory therefore does not constitute persistent agent memory. The no-memory ablation removes the former while leaving each detector implementation unchanged.

\subsection{Model bank}

The bank contains exactly eleven candidates. Nine continuous-time dynamic graph encoders span recurrent state, event processes, temporal attention, anonymous walks, sequence mixing, and Transformer histories. We adapt their anomaly-scoring implementations from BAG \citep{hua2026bag}, while attributing each architecture to its original paper in the candidate registry. GTAN combines transaction attributes with graph context \citep{xiang2023gtan}; THG-OAFN combines temporal, heterogeneous-graph, and imbalance-aware components \citep{wei2025thgoafn}. The two fusion models extend the bank beyond temporal interaction encoders.

Fusion and temporal models receive the same label-free source attributes; adapters may add train-history-only structural controls required by their native interfaces. For edge tasks, each transaction is lifted to a prediction node while its sender, receiver, and chronological relations define the graph. For node tasks, the native prediction unit is retained. This adapter choice lets all eleven candidates be selectable on every dataset without changing the anomaly unit.

\section{Benchmark Construction}

\subsection{Datasets}

Table~\ref{tab:datasets} covers three node-level and four edge-level tasks. DGraph-Fin and Elliptic++ provide two additional node-level financial anomaly tasks \citep{huang2022dgraph,elmougy2023ellipticpp}. MOOC is the temporal interaction stream distributed by BAG and provides a non-financial anomaly environment that broadens the benchmark beyond transaction graphs \citep{hua2026bag}. Bitcoin-OTC is the signed Bitcoin trust network used by BAG; following its anomaly protocol, we replace a fixed seeded 5\% of chronological interactions with cross-community non-edges and keep the same injected labels throughout the rolling history. IEEE-CIS provides timestamped transaction fraud labels and rich attributes; we construct sender/receiver-style entity links from stable categorical identities. The competition files require separate acceptance of the upstream access terms.\footnote{\url{https://www.kaggle.com/competitions/ieee-fraud-detection/data}} IBM HI-Small is a large synthetic AML transaction stream with account-level interaction structure; we use its transaction label as the supervised edge unit \citep{altman2023aml}. S-FFSD is the public simulated transaction stream distributed with the GTAN implementation \citep{xiang2023gtan}.

\begin{table}[t]
\centering
\caption{Supervised prediction units used to construct the benchmark. Each non-bootstrap prediction unit appears in exactly one hidden deployment window.}
\label{tab:datasets}
\small
\begin{tabular}{lrrrl}
\toprule
Dataset & Units & Positives & Rate & Level \\
\midrule
DGraph-Fin & 964,840 & 7,503 & 0.78\% & node \\ Elliptic++ & 46,564 & 4,545 & 9.76\% & node \\ MOOC & 411,749 & 4,066 & 0.99\% & node \\
Bitcoin-OTC & 35,592 & 1,780 & 5.00\% & edge \\
IEEE-CIS & 590,540 & 20,663 & 3.50\% & edge \\
IBM HI-Small & 5,078,345 & 5,177 & 0.10\% & edge \\
S-FFSD & 29,643 & 5,256 & 17.73\% & edge \\
\bottomrule
\end{tabular}
\end{table}

The resulting environments exhibit different kinds of temporal variation rather than merely repeating one static split. Across the eight deployment windows, positive prevalence ranges from 0.26--1.00\% on DGraph-Fin, 3.25--23.50\% on Elliptic++, 0.73--1.25\% on MOOC, 4.50--5.54\% on Bitcoin-OTC, 3.10--4.31\% on IEEE-CIS, 0.05--0.69\% on IBM HI-Small, and 6.68--41.36\% on S-FFSD. Window sizes range from 33,586--176,818, 2,775--7,458, 40,459--42,752, 3,559--3,560, 58,818--59,512, 148,560--881,476, and 2,964--2,965 prediction units, respectively. Reporting both AP and fixed-depth capture is therefore necessary: AP's baseline changes with prevalence, whereas capture measures how much anomaly or fraud a fixed review queue recovers.

\subsection{Eight windows and delayed labels}

We sort supervised prediction units by timestamp without splitting equal-time events. The earliest 20\% forms a bootstrap history; 70\% of its distinct timestamps provide initial training history and the remainder provides the first mature validation interval. The remaining timestamps are partitioned into exactly eight contiguous groups. At $e_0$ and $e_1$, the bootstrap partitions remain the only mature supervision. At $e_k$ for $k\geq2$, $W_{k-2}$ becomes validation evidence and all earlier mature windows enter training. Every $W_k$ is scored exactly once.

Eight windows provide six decisions with matured deployment feedback ($e_2$--$e_7$) while retaining enough positives per window for meaningful fixed-depth capture on the most imbalanced stream. The manifests fix timestamp boundaries rather than equal row counts, so deployment volume is allowed to drift.

This construction distinguishes three data roles. \emph{Training} contains mature historical labels used for optimization. \emph{Validation} is the newest mature interval used for early stopping and later feedback. \emph{Deployment} is the current hidden window: its labels are unavailable to the agent and excluded from optimization and early-stopping decisions; the evaluator reads them only for final scoring. We use ``deployment'' instead of ``test'' in all agent-facing schemas to reduce accidental field confusion, while the payload guard forbids either evaluator term.

Evaluation is batched within a deployment window. After the controller commits and training finishes, the detector may use the window's unlabeled attributes and topology to produce its review ranking, but neither its labels nor candidate metrics. This permits graph--attribute fusion models with snapshot-style inference; the benchmark does not claim per-event online latency or strict within-window topology causality.

\subsection{Training and evaluation}

Every selected detector is initialized afresh in each episode and trained on all mature history. Models use an NVIDIA A10G GPU for at most 20 epochs with validation-AP early stopping, patience three, and seed 42. Batch size is 1024 for Bitcoin-OTC, IEEE-CIS, and S-FFSD, 2048 for MOOC, and 4096 for IBM HI-Small. IBM HI-Small uses a three-epoch cap because its mature history grows to millions of transactions; the other datasets use a 20-epoch cap. No examples are subsampled. BAG models use 32-dimensional memory/features, one temporal layer, two attention heads where applicable, ten historical neighbors, dropout 0.1, and learning rate $10^{-3}$. Numeric transforms are fitted independently in each episode using only units visible in mature training history and then applied unchanged to validation and deployment records. Fusion adapters restore the validation-best state; the pinned upstream BAG node loop evaluates its terminal early-stopped state. After training terminates, the resulting model is evaluated on the deployment window once.

AP is the primary metric because anomaly rates vary substantially. For operational interpretation, $\capmetric{q}$ is the fraction of all positives captured among the top $q\%$ scores; we report $q\in\{3,5\}$. AUROC is secondary. The benchmark's primary leaderboard score is the equal-weight macro average of each dataset's mean AP across its eight deployment windows; $\capmetric{5\%}$ breaks a numerical tie. A submission is rankable only if all eight decisions per dataset pass the registry, timestamp, payload-leakage, declared-training-scope, and single-hidden-evaluation audits. We additionally record (i) architecture switches, (ii) legal-choice and strict-JSON rates, (iii) contract-normalized actions and malformed-response recoveries, (iv) whether a switch follows a matured AP decline, (v) validation-to-deployment AP gap, and (vi) distinct trained candidates, language-model tokens, and GPU-job wall-clock time. A controller is charged the elapsed time of each candidate it selects, even when two controllers share one executed artifact; this makes each row the standalone cost of that policy rather than giving accidental credit for another controller's matching choice. Dataset-level macro averages prevent larger streams from dominating smaller ones, while complete trajectories remain mandatory so that the aggregate score cannot hide pathological windows.

\section{Benchmark Validation Design}

\paragraph{Controllers.}
The two primary controllers are OpenAI GPT-6 Astra and Anthropic Claude Opus 5, both invoked through the same managed inference API with identical instructions and maximum output length. A separate expanded sweep configures Amazon Nova Pro v1, Mistral Large 3 675B Instruct, and DeepSeek V3.2; we analyze datasets for which all eight deployment windows are complete. Temperature is not set because not every hosted model supports that field. Each controller has a dataset-specific memory namespace; selections, rationales, and model state never transfer across datasets. A malformed reply may be recovered only by extracting an explicit registered candidate identifier and action; if none is recoverable, the executor retries the same decision at most twice with a format reminder. It retains every raw reply and token count. No current-window training begins until a legal choice is persisted.

\paragraph{No-memory reference.}
We rerun the GPT-6 Astra controller with both \(a_{k-1}\) and \(H_{\leq k-2}\) removed. It still observes the current aggregate dataset profile and identical model cards. Its selection is persisted before a fresh training run. Comparing this condition with the stateful controller tests the specific value of outcome memory without exposing current candidate metrics.

\paragraph{Statistical analysis.}
The basic unit is a paired dataset--window outcome. We report complete trajectories and per-dataset means, with an equal-weight macro average across the four datasets shared by the primary and no-memory controllers. The additional-controller comparison reports per-dataset means on three shared datasets. Confidence intervals use 10,000 moving-block bootstrap replicates with block length two within each dataset, followed by equal-weight aggregation across datasets (seed 42); stateful-versus-no-memory comparisons reuse the same sampled blocks. With one training seed and four evaluated primary-controller datasets, we interpret comparisons descriptively, emphasizing effect direction and cross-dataset consistency rather than a single pooled $p$-value.

\section{Results}

\subsection{Hidden-window detection}

Table~\ref{tab:main-results} reports deployment metrics from four datasets with complete eight-window primary-controller trajectories; no retrospective train-all result is included. GPT-6 Astra, Claude Opus 5, and the GPT-6 Astra no-memory reference share the same eight hidden windows and candidate bank. Fixed-depth capture prevents a high AP caused by changing class prevalence from being mistaken for broad fraud recovery.

\begin{table*}[t]
\centering
\caption{Four-dataset agent-first results on hidden deployment windows. Per-dataset detection metrics average eight chronological windows; bold marks the best controller in each dataset and metric. The macro row averages dataset means, while switches and GPU hours sum across datasets. NM denotes the GPT-6 Astra condition without outcome memory; C@3 and C@5 are positive capture at fixed 3\% and 5\% review budgets. Shared training is charged to each selecting policy.}
\label{tab:main-results}
\small
\begin{tabular}{llrrrrrr}
\toprule
& & \multicolumn{4}{c}{Detection quality} & \multicolumn{2}{c}{Behavior and cost} \\
\cmidrule(lr){3-6}\cmidrule(l){7-8}
Dataset & Ctrl. & AP & AUC & C@3 & C@5 & Switches & GPU h \\
\midrule
DGraph-Fin & GPT-6 Astra & 0.0069 & 0.5375 & 0.0344 & 0.0545 & 2 & 0.02 \tabularnewline
DGraph-Fin & Claude Opus 5 & 0.0070 & 0.5424 & 0.0382 & 0.0594 & 2 & 0.38 \tabularnewline
DGraph-Fin & GPT-6 Astra (NM) & 0.0072 & 0.5560 & 0.0415 & 0.0659 & 0 & 0.02 \tabularnewline
\midrule
Elliptic++ & GPT-6 Astra & 0.1720 & 0.6170 & 0.0238 & 0.0430 & 1 & 0.03 \tabularnewline
Elliptic++ & Claude Opus 5 & 0.1436 & 0.5701 & 0.0085 & 0.0175 & 4 & 0.01 \tabularnewline
Elliptic++ & GPT-6 Astra (NM) & 0.1534 & 0.5852 & 0.0162 & 0.0329 & 0 & 0.00 \tabularnewline
\midrule
IEEE-CIS & GPT-6 Astra & 0.0469 & 0.5538 & 0.0464 & 0.0728 & 1 & 0.47 \tabularnewline
IEEE-CIS & Claude Opus 5 & 0.1427 & 0.6913 & 0.1505 & 0.2045 & 1 & 3.48 \tabularnewline
IEEE-CIS & GPT-6 Astra (NM) & 0.0446 & 0.5351 & 0.0411 & 0.0692 & 0 & 0.01 \tabularnewline
\midrule
S-FFSD & GPT-6 Astra & 0.3100 & 0.6665 & 0.0522 & 0.0864 & 2 & 0.04 \tabularnewline
S-FFSD & Claude Opus 5 & 0.3722 & 0.7503 & 0.0810 & 0.1392 & 1 & 0.04 \tabularnewline
S-FFSD & GPT-6 Astra (NM) & 0.3097 & 0.6437 & 0.0458 & 0.0728 & 0 & 0.00 \tabularnewline
\midrule
Macro & GPT-6 Astra & 0.1340 & 0.5937 & 0.0392 & 0.0642 & 6 & 0.56 \tabularnewline
Macro & Claude Opus 5 & 0.1664 & 0.6385 & 0.0695 & 0.1051 & 8 & 3.91 \tabularnewline
Macro & GPT-6 Astra (NM) & 0.1287 & 0.5800 & 0.0362 & 0.0602 & 0 & 0.03 \tabularnewline
\bottomrule
\end{tabular}
\end{table*}

We compare the three policies within each dataset and report their equal-weight macro averages in Table~\ref{tab:main-results}. The fixed-depth capture columns complement AP when positive prevalence shifts across windows. The selected GPU-hours column counts only architectures the policy actually requested; it does not hide a train-all evaluator behind the controller. IEEE-CIS and S-FFSD include a separately declared fixed TGN diagnostic, excluded from this policy cost.

\IfFileExists{figures/rolling_trajectories.pdf}{%
\begin{figure*}[t]
\centering
\includegraphics[width=0.92\textwidth]{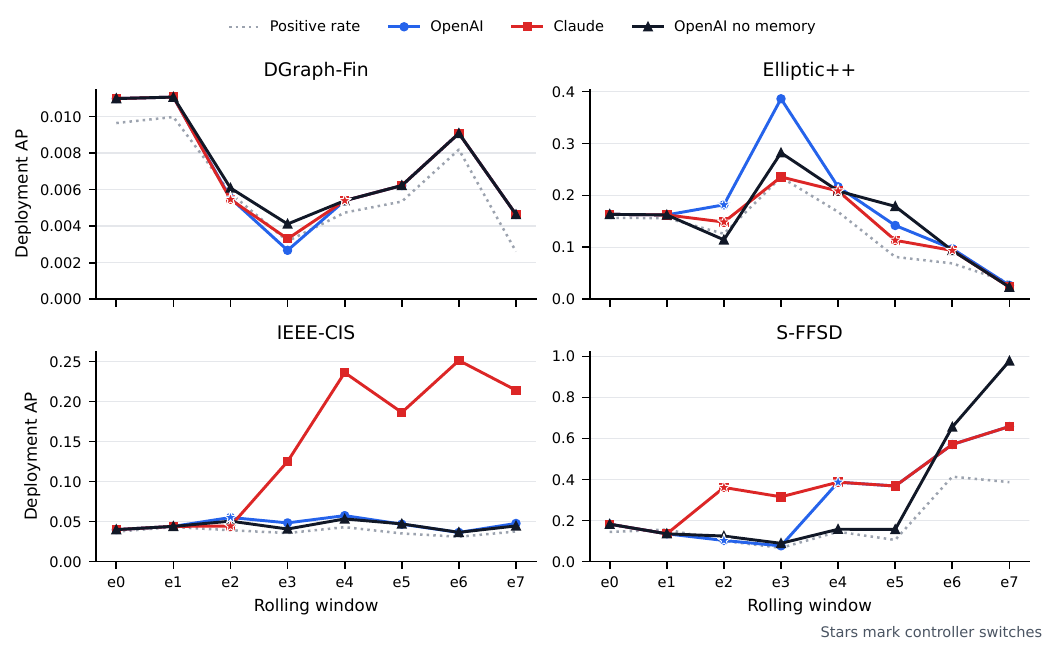}
\caption{Four-dataset per-window deployment AP, controller switches (stars), and positive prevalence. Prevalence is the expected AP of random ranking, so the dotted line helps distinguish improved ranking from an easier class balance. Every controller choice precedes current-window training and scoring.}
\label{fig:rolling-trajectories}
\end{figure*}
}{}

\subsection{Additional controllers on completed datasets}\label{sec:additional-controllers} The three additional controllers---Amazon Nova Pro v1, Mistral Large 3 675B Instruct, and DeepSeek V3.2---have completed all eight windows on MOOC, Bitcoin-OTC, and S-FFSD under the same candidate bank and one-window label delay. Table~\ref{tab:additional-controller-results} reports the mean hidden-window scores from these run-complete trajectories. Each controller committed to its candidate before current-window training. \par \begin{table}[t]\centering\caption{Completed additional-controller runs on three datasets. AP, AUROC, and C@3 are unweighted means over eight chronological hidden deployment windows; C@3 is positive capture at a 3\% review budget.}\label{tab:additional-controller-results}\small\begin{tabular}{llrrr}\toprule Dataset & Controller model & AP & AUROC & C@3 \\ \midrule MOOC & Amazon Nova Pro v1 & 0.0146 & 0.5934 & 0.0656 \\ & Mistral Large 3 675B Instruct & 0.0143 & 0.6209 & 0.0499 \\ & DeepSeek V3.2 & 0.0141 & 0.6156 & 0.0491 \\ \midrule Bitcoin-OTC & Amazon Nova Pro v1 & 0.1596 & 0.7501 & 0.1335 \\ & Mistral Large 3 675B Instruct & 0.1648 & 0.5748 & 0.1395 \\ & DeepSeek V3.2 & 0.2520 & 0.6349 & 0.2284 \\ \midrule S-FFSD & Amazon Nova Pro v1 & 0.3793 & 0.7792 & 0.0897 \\ & Mistral Large 3 675B Instruct & 0.4261 & 0.8180 & 0.1013 \\ & DeepSeek V3.2 & 0.3570 & 0.7559 & 0.0648 \\ \bottomrule\end{tabular}\end{table} The three completed streams expose different limits of controller selection. MOOC has an eight-window mean anomaly rate of 0.0096, the expected AP of random ranking. The three additional controllers obtain mean AP between 0.0141 and 0.0146 and C@3 between 0.0491 and 0.0656. Their AUROC values of 0.5934--0.6209 are more separated than their AP values. Thus the observed selection policies yield only modest positive-class ranking gains on MOOC, even though AUROC is above 0.5; the experiment does not establish how an unselected detector would have performed. \par Bitcoin-OTC has a higher mean anomaly rate of 0.0498 and much greater variation between selection paths. Among the additional controllers, DeepSeek V3.2 reaches AP 0.2520, compared with 0.1596 for Amazon Nova Pro v1 and 0.1648 for Mistral Large 3. The completed primary Claude Opus 5 trajectory reaches AP 0.5227 and selects GraphMixer in six of eight windows. Its GraphMixer windows have AP 0.595--0.728, whereas its two DyGFormer windows have AP 0.071 and 0.154. The temporary drop and recovery coincide with the model switches; they are not a controlled estimate of the switch effect because the unselected alternative was not trained in those same windows. \par The choice of evaluation metric changes the controller ranking. On Bitcoin-OTC, Amazon Nova Pro v1 has higher AUROC than DeepSeek V3.2 (0.7501 versus 0.6349), but lower AP (0.1596 versus 0.2520) and C@3 (0.1335 versus 0.2284). For a fixed review budget, AUROC alone would therefore favor the controller that captures fewer anomalies near the top of the queue. More frequent switching is also not sufficient for improvement: Mistral Large 3 switches seven times on Bitcoin-OTC and obtains AP 0.1648, while DeepSeek V3.2 switches five times and obtains AP 0.2520. These are descriptive associations, not causal effects of switch count. \par S-FFSD supplies a third outcome regime. Its mean anomaly rate is 0.1898, rising to 0.4136 and 0.3870 in the last two windows, so aggregate AP must be read alongside the changing base rate. Mistral Large 3 675B Instruct reaches mean AP 0.4261 and C@3 0.1013, compared with 0.3793 and 0.0897 for Amazon Nova Pro v1 and 0.3570 and 0.0648 for DeepSeek V3.2. The Mistral--DeepSeek AP gap is concentrated in windows e0002 and e0003: DeepSeek selects GTAN and obtains AP 0.124 and 0.089, whereas Mistral selects TGN and DyGFormer and obtains AP 0.361 and 0.336. These are same-window comparisons among the candidates selected by the controllers; they do not reveal the scores of untrained alternatives. The S-FFSD decision files also pass the source-node timestamp audit for decision-before-training order. \par Delayed outcome memory does not provide a uniform gain in the completed original-controller runs. GPT-6 Astra without cross-window memory exceeds its stateful counterpart on MOOC (AP 0.0152 versus 0.0140) and Bitcoin-OTC (0.2637 versus 0.1220); the stateful counterpart is only slightly higher on IEEE-CIS (0.0469 versus 0.0446) and S-FFSD (0.3100 versus 0.3097). The observed value of feedback depends on the decisions it changes, rather than the presence of memory alone. \par \paragraph{Controller capabilities revealed by the trajectories.} The action-level record distinguishes a legal decision from a useful one. On Bitcoin-OTC, Claude Opus 5 selects GraphMixer in six windows while other controllers follow lower-AP paths; on S-FFSD, Mistral Large 3 675B Instruct selects TGN and DyGFormer in two low-prevalence windows where DeepSeek V3.2 selects GTAN and scores lower. These observations associate backbone choice under delayed feedback with substantially different hidden-window ranking, but the missing scores of unselected candidates prevent causal attribution to any single choice. The AUROC/AP reversal on Bitcoin-OTC shows why realized choices must be judged against rare-positive ranking and fixed-budget capture. The MOOC plateau, the high-switching low-AP Bitcoin-OTC trajectory, and the no-memory wins show that tool access, frequent switching, and retaining feedback are not sufficient on their own. The benchmark measures whether a controller can turn mature evidence into a legal pre-training commitment that improves later AP and capture; these outcomes do not establish a controller's internal reasoning. \par \subsection{Adaptation behavior}

A controller can obtain acceptable mean AP for the wrong reason---for example, by never changing a generally strong architecture. We therefore inspect every transition\IfFileExists{figures/rolling_trajectories.pdf}{ in Figure~\ref{fig:rolling-trajectories}}{} and in Appendix Table~\ref{tab:decision-results}. The table separates persistence from evidence-responsive switching and reports the deployment change following each matured negative outcome. Rationales are used only as diagnostics; correctness is determined by legal actions and later deployment scores.
Appendix Table~\ref{tab:decision-results} reports protocol validity and controller behavior separately from detection quality. A legal model choice can still be a poor operational decision: switching consumes training time and can reduce later hidden-window AP. Conversely, a controller can improve mean AP without responding to newly matured negative evidence. The trajectories expose these distinctions episode by episode.

\subsection{Does delayed memory help?}

Table~\ref{tab:ablation-results} compares the stateful GPT-6 Astra controller with the no-memory ablation. Useful memory should improve future AP or capture without an excessive rise in switching and compute. If choices remain identical, the correct conclusion is that this prompt and candidate metadata do not cause the agent to use the delayed outcomes, not that memory is intrinsically unnecessary.

\begin{table}[t]
\centering
\caption{GPT-6 Astra cross-window memory ablation, macro-averaged over the four evaluated datasets.}
\label{tab:ablation-results}
\small
\begin{tabular}{lrrrrr}
\toprule
Condition & AP & $\capmetric{3\%}$ & $\capmetric{5\%}$ & Switches & GPU h \\
\midrule
GPT-6 Astra & 0.1340 & 0.0392 & 0.0642 & 6 & 0.56 \tabularnewline
GPT-6 Astra no memory & 0.1287 & 0.0362 & 0.0602 & 0 & 0.03 \tabularnewline
\bottomrule
\end{tabular}
\end{table}

The no-memory reference isolates the effect of exposing prior selections and matured outcomes in the controller payload. We compare paired deployment windows and report switching and selected training time alongside macro AP. This distinction matters because a memory-enabled controller may react to feedback yet fail to improve ranking or may improve ranking only by choosing a substantially more expensive architecture.

\section{Related Work} Dynamic and graph anomaly detectors model structural, attribute, and temporal deviations, including fraud-specific graph context \citep{dou2020caregnn,liu2022cola,rossi2020tgn,yu2023dygformer,hua2026bag}. Agent benchmarks evaluate tool use and machine-learning experimentation \citep{jimenez2024swebench,mialon2023gaia,huang2024mlagentbench,chan2024mlebench}, while graph-learning agents configure models and workflows \citep{wei2024glagent,zheng2025llmnet,signgad2026}. \bench{} evaluates the decision to select a registered dynamic-graph detector before current-window training and assesses that choice only after future labels mature. Appendix~\ref{app:related-work} details the closest model and agent families. \section{Conclusion} We introduce \bench{}, an executable benchmark for agent-first lifecycle control in dynamic graph anomaly detection under delayed labels. Its fixed candidate registry, typed actions, time-causal executor, and decision-before-training audit make each controller choice observable and its later deployment outcome comparable.\section*{Reproducibility Statement}
An anonymous code artifact is available at \url{https://anonymous.4open.science/r/DGB-2027-8F3C/}. It contains the rolling manifests, graph preparation and training adapters, the eleven-model registry, the controller executor, and aggregation code. The executor records each decision before training and retains artifacts for auditing decision timing, training scope, and hidden-window evaluation. Raw datasets, credentials, checkpoints, and run outputs are not part of the code artifact; the README lists data sources, access conditions, and pinned upstream dependencies.

\section*{Ethics Statement}
Fraud scores can cause investigation burden and disparate false positives. This work evaluates public or simulated research data and does not claim deployment readiness. We therefore report fixed-budget capture together with aggregate ranking metrics and discourage interpreting graph proximity as guilt. The agent receives no raw personal identifiers. IEEE-CIS is used under its competition terms and is not redistributed.

\section*{AI Use Statement}
Generative AI tools were used for methodology refinement, experiment-code generation and debugging, literature discovery, writing and editing, table preparation, scientific-figure scripting, and assistance with result interpretation. Human authors selected the research question, reviewed the causal protocol, inspected source repositories and official policies, executed and tested the code, and will verify every reported number against saved artifacts. All AI-assisted text and code were reviewed by the authors, who take responsibility for the final content, claims, and artifacts.

\bibliography{references}
\bibliographystyle{iclr2027_conference}

\appendix \section{Detailed Controller Behavior}\label{app:behavior} \begin{table*}[t]\centering\caption{Four-dataset controller behavior over eight windows per dataset. JSON is the accepted response's strict-parse rate; Retry counts extra format-recovery calls. A feedback response is a keep/switch decision immediately after a newly matured AP decline becomes visible. $\Delta$AP compares the following deployment outcome with the preceding one and is descriptive, not causal.}\label{tab:decision-results}\small\begin{tabular}{llrrrrrrr}\toprule& & \multicolumn{3}{c}{Protocol} & \multicolumn{4}{c}{Adaptation} \\\cmidrule(lr){3-5}\cmidrule(l){6-9}Dataset & Ctrl. & Legal & JSON & Retry & Models & Switch & Resp. & $\Delta$AP \\\midrule DGraph-Fin & GPT-6 Astra & 1.00 & 1.00 & 2 & 2 & 1/2 & -0.0014 \tabularnewline
DGraph-Fin & Claude Opus 5 & 1.00 & 0.12 & 2 & 2 & 1/2 & -0.0018 \tabularnewline
Elliptic++ & GPT-6 Astra & 1.00 & 1.00 & 2 & 1 & 0/3 & 0.0194 \tabularnewline
Elliptic++ & Claude Opus 5 & 1.00 & 0.12 & 2 & 4 & 2/4 & -0.0388 \tabularnewline
IEEE-CIS & GPT-6 Astra & 1.00 & 1.00 & 2 & 1 & 0/2 & 0.0111 \tabularnewline
IEEE-CIS & Claude Opus 5 & 1.00 & 0.25 & 2 & 1 & 0/1 & 0.0004 \tabularnewline
S-FFSD & GPT-6 Astra & 1.00 & 1.00 & 3 & 2 & 1/4 & 0.1390 \tabularnewline
S-FFSD & Claude Opus 5 & 1.00 & 0.25 & 2 & 1 & 0/3 & 0.2258 \tabularnewline
\bottomrule
\end{tabular}\end{table*}  \section{Detailed Additional Controller Results}\label{app:additional-controller-details} \par The three completed streams expose different limits of controller selection. MOOC has an eight-window mean anomaly rate of 0.0096, the expected AP of random ranking. The three additional controllers obtain mean AP between 0.0141 and 0.0146 and C@3 between 0.0491 and 0.0656. Their AUROC values of 0.5934--0.6209 are more separated than their AP values. Thus the observed selection policies yield only modest positive-class ranking gains on MOOC, even though AUROC is above 0.5; the experiment does not establish how an unselected detector would have performed. \par Bitcoin-OTC has a higher mean anomaly rate of 0.0498 and much greater variation between selection paths. Among the additional controllers, DeepSeek V3.2 reaches AP 0.2520, compared with 0.1596 for Amazon Nova Pro v1 and 0.1648 for Mistral Large 3. The completed primary Claude Opus 5 trajectory reaches AP 0.5227 and selects GraphMixer in six of eight windows. Its GraphMixer windows have AP 0.595--0.728, whereas its two DyGFormer windows have AP 0.071 and 0.154. The temporary drop and recovery coincide with the model switches; they are not a controlled estimate of the switch effect because the unselected alternative was not trained in those same windows. \par The choice of evaluation metric changes the controller ranking. On Bitcoin-OTC, Amazon Nova Pro v1 has higher AUROC than DeepSeek V3.2 (0.7501 versus 0.6349), but lower AP (0.1596 versus 0.2520) and C@3 (0.1335 versus 0.2284). For a fixed review budget, AUROC alone would therefore favor the controller that captures fewer anomalies near the top of the queue. More frequent switching is also not sufficient for improvement: Mistral Large 3 switches seven times on Bitcoin-OTC and obtains AP 0.1648, while DeepSeek V3.2 switches five times and obtains AP 0.2520. These are descriptive associations, not causal effects of switch count. \par S-FFSD supplies a third outcome regime. Its mean anomaly rate is 0.1898, rising to 0.4136 and 0.3870 in the last two windows, so aggregate AP must be read alongside the changing base rate. Mistral Large 3 675B Instruct reaches mean AP 0.4261 and C@3 0.1013, compared with 0.3793 and 0.0897 for Amazon Nova Pro v1 and 0.3570 and 0.0648 for DeepSeek V3.2. The Mistral--DeepSeek AP gap is concentrated in windows e0002 and e0003: DeepSeek selects GTAN and obtains AP 0.124 and 0.089, whereas Mistral selects TGN and DyGFormer and obtains AP 0.361 and 0.336. These are same-window comparisons among the candidates selected by the controllers; they do not reveal the scores of untrained alternatives. The S-FFSD decision files also pass the source-node timestamp audit for decision-before-training order. \par \section{Extended Related Work}\label{app:related-work} \paragraph{Dynamic and graph anomaly detection.}One line of work improves graph fraud detectors at a fixed point in time by handling camouflage, imbalance, relation semantics, and temporal structure \citep{dou2020caregnn,liu2021pcgnn,zhuo2024pmp,huang2022dgraph}. Another models attribute and structural deviations \citep{ding2019dominant,liu2022cola,xu2022conad,roy2024gadnr} or learns directly from event sequences \citep{rossi2020tgn,yu2023dygformer}. \bench{} builds on this work: temporal and graph--attribute detectors are selectable tools in the benchmark. The additional question is whether an agent can choose among them as graphs, entities, and labels change, using only feedback available at decision time.\paragraph{Agents for tool use and autonomous model building.}A second line of work evaluates agents through executed actions rather than generated text. SWE-bench, GAIA, BFCL, and tau-bench test software repair, tool use, function calling, and interactive reliability \citep{jimenez2024swebench,mialon2023gaia,patil2025bfcl,yao2024taubench}. MLAgentBench, MLE-bench, the AI Scientist, and AIBuildAI study iterative experimentation, autonomous model engineering, and collaborative build--test loops \citep{huang2024mlagentbench,chan2024mlebench,lu2024aiscientist,aibuildai2026}. We treat hierarchical roles and iterative validation as established baseline capabilities. Cost-aware agent leaderboards and planning benchmarks further show why utility must be reported with resource use \citep{kapoor2026hal,liu2025costbench}; accordingly, we report selected detector-training time alongside detection quality. These benchmarks do not impose the particular sequence in which a graph detector is selected before training, deployed on a hidden window, and assessed only after its labels mature.\paragraph{Agents for graph learning.}Closer to our setting, GL-Agent and LLMNet automate graph-learning configuration \citep{wei2024glagent,zheng2025llmnet}, while LLaGA and Think-on-Graph connect language models to graph representations and traversal \citep{chen2024llaga,sun2024thinkgraph}. SignGAD is particularly close: its planner uses a task description and graph statistics to build candidate topology, evidence, and detector workflows, then selects a workflow through validation search with a guarded final refit \citep{signgad2026}. SignGAD asks whether an agent can design a better workflow for a static graph and fixed label budget. \bench{} asks how a controller chooses a registered detector repeatedly across chronological windows, committing before current-window training and using only outcomes that have matured. The contribution is the delayed-feedback selection protocol and its executable evaluation, rather than a new detector layer. 

\section{Per-Window Information Contract}
\label{app:contract}

The agent payload contains dataset identifier, task level, episode identifier, a one-window-delay declaration, mature train/validation row and positive counts, current unlabeled row count, eleven model cards, previous selection, matured self-history, and a sandbox declaration. It excludes current positives, all current model metrics, evaluator paths, raw files, shell commands, and other-agent traces. Unit tests recursively reject forbidden evaluator keys before any model request is issued.

\section{Benchmark Positioning} \par BAG evaluates dynamic graph anomaly detectors across ten datasets and supplies a unified implementation and scoring interface \citep{hua2026bag}. The Temporal Graph Benchmark (TGB) standardizes temporal graph tasks and chronological evaluation \citep{huang2023tgb}, while BenchTemp compares temporal graph neural networks under fixed datasets and evaluation pipelines \citep{huang2024benchtemp}. These benchmarks assess predictive models after their training protocol has been specified. Their published tasks do not require an agent to commit to one architecture before the current window is trained and scored. \par MLAgentBench evaluates agents carrying out machine-learning experimentation on thirteen tasks \citep{huang2024mlagentbench}; MLE-bench evaluates machine-learning engineering agents on seventy-five offline competition tasks \citep{chan2024mlebench}. Both test broader agent execution, but their task units are fixed problems rather than a rolling stream with delayed labels. DynGraphAgentBench instead defines controller episodes over seven environments and eight chronological windows per environment. The choice is recorded before current-window training, the hidden outcome is released only after a one-window delay, and the verifier checks that this order was respected. Thus the evaluand is sequential model choice under feedback and compute constraints.

\section{Candidate Registry} \par The bank contains eleven selectable candidates. The nine continuous-time encoders use the anomaly-detection implementation supplied by BAG \citep{hua2026bag}; their architectures and scientific provenance are described by the original model papers below. The two remaining candidates are fraud-oriented graph--attribute models. All eleven receive the same legal pre-training information, and no tabular-only or static-graph candidate is selectable. The mechanisms below are selection cues, not claims that a model is optimal on any benchmark window. \par JODIE couples recurrent updates of interacting entities with a projection of their embeddings forward in time, making it a candidate when interaction order and time since the last event are informative \citep{kumar2019jodie}. DyRep explicitly models topological change and node interactions on two time scales, testing whether those event processes should be represented jointly \citep{trivedi2019dyrep}. TGN maintains per-node memory updated by temporal messages and then computes event-time embeddings, offering an explicit persistent state within a training episode \citep{rossi2020tgn}. These are distinct memory mechanisms; detector state is reset between benchmark windows and is never the controller memory. \par TGAT uses functional time encodings and attention over past temporal neighbors, so its representation can attend to history without maintaining a recurrent node state \citep{xu2020tgat}. TCL encodes the two endpoints through separate temporal streams and co-attention, with a contrastive objective in its original formulation; it tests whether coupled endpoint histories add useful interaction context \citep{wang2021tcl}. CAWN samples time-respecting anonymous walks to encode temporal motifs while reducing dependence on fixed node identities, a relevant inductive bias when entities change across windows \citep{wang2021cawn}. \par GraphMixer combines an MLP link encoder with mean-pooled neighbor information, providing a simpler temporal-neighborhood alternative to recurrent and attention-heavy encoders \citep{cong2023graphmixer}. DyGFormer patches long interaction histories for a Transformer and explicitly encodes neighbor co-occurrence between endpoints, testing the value of long-range sequence and pairwise-history signals \citep{yu2023dygformer}. FreeDyG encodes interaction frequencies and shared-neighbor frequency patterns, targeting repeated or shifting interaction behavior \citep{tian2024freedyg}. \par GTAN builds an attribute-driven graph representation with attention for transaction fraud, testing whether feature-rich transaction attributes and graph context are complementary \citep{xiang2023gtan}. THG-OAFN combines temporal GRU and graph branches with heterogeneous-relation attention and graph oversampling for imbalanced fraud data \citep{wei2025thgoafn}. The benchmark adapters make these two fraud-oriented architectures available under the same decision contract as the temporal encoders; their original training objectives and datasets should not be read as evidence of superiority on the seven benchmark environments.

\section{Audit Checks}

The formal runner asserts exactly eight windows, one-window label delay, eleven model cards, a legal candidate identifier, and a pre-training decision marker. For node tasks, the BAG runner is called with one model name; for edge tasks, one candidate directory is created per distinct selection. Deployment loaders are never used for early-stopping decisions. The run summary explicitly records \texttt{selection\_before\_training=true}, the declared reference condition, and the candidates actually trained. These checks prevent a resume operation from silently reintroducing train-all selection.

\section{Agent Prompt and Execution Details}\label{app:agent_prompt} \paragraph{Decision point and model interface.} The formal controller is invoked once per dataset and window, before any detector selected for that window is trained. The experiment has eight chronological windows per dataset and a fixed delay of one complete window for deployment labels. Each invocation is an independent Amazon Bedrock \texttt{Converse} call to the controller model under study. The runner sends a single user message consisting of the fixed instruction below followed immediately by a JSON serialization of the window payload with keys sorted. It does not supply a separate system message, exemplars, a chain-of-thought template, or tool definitions. The response cap is \texttt{maxTokens=1200}; sampling parameters are left at the provider defaults. The same instruction and payload construction are used for GPT-6 Astra, Claude Opus 5, Amazon Nova Pro v1, Mistral Large 3 675B Instruct, and DeepSeek V3.2, so the controller comparison changes the hosted model rather than the task wording. \par \paragraph{Fixed instruction.} The following is the complete static instruction preceding the serialized payload in the formal rolling runner (LaTeX formatting of field names does not change the text sent to the models): \begin{quote}\small You control model selection for a causal rolling fraud-detection experiment. Choose exactly one \texttt{candidate\_id} from \texttt{available\_candidates} BEFORE that candidate is trained in the current window. You may use the dataset profile, model cards, the previous choice, and only outcomes whose labels have already matured. Current deployment labels and all current candidate metrics are unavailable. Return exactly one JSON object with \texttt{candidate\_id}, \texttt{action} (\texttt{initialize}, \texttt{keep}, or \texttt{switch}), \texttt{rationale}, and \texttt{evidence}. If \texttt{previous\_selection} is null, action must be \texttt{initialize}; otherwise use \texttt{keep} when \texttt{candidate\_id} is unchanged and \texttt{switch} when it changes. Do not invent or combine models. \end{quote} \par \paragraph{Payload contents and information boundary.} The appended JSON identifies the dataset, node- or edge-level task, episode identifier, and the one-window label delay. Its \texttt{dataset\_profile} reports counts of mature training rows and positives, mature validation rows and positives, and current unlabeled rows. The \texttt{available\_candidates} array has exactly eleven model cards. Each card supplies a registered identifier, model name, family, task level, short architectural description, and the status \texttt{not\_yet\_trained\_in\_current\_window}; it supplies no current score. The payload also contains \texttt{previous\_selection}, \texttt{memory\_mode}, and \texttt{matured\_history}. Each history item records the earlier window, selected candidate, action, validation metrics, and deployment metrics whose labels have matured. At decision window $W_k$, the newest permitted deployment outcome is $W_{k-2}$; thus the history is empty for $W_0$ and $W_1$, although $W_1$ can still include the previous choice. The \texttt{sandbox\_contract} explicitly marks filesystem access, shell access, network tools, cross-agent state, current-window labels, and current-window metrics as unavailable. Before the Bedrock call, a recursive guard rejects payload keys containing \texttt{test} or \texttt{oracle}. The agent sees neither the evaluator's candidate metrics file nor the hidden labels. \par  \paragraph{Response validation and retries.} The runner requests one JSON object with a registered \texttt{candidate\_id} and an action label. It first parses a JSON object, allowing a surrounding Markdown code fence. If malformed prose prevents full JSON parsing, a restricted recovery path extracts only \texttt{candidate\_id} and \texttt{action}; the original response is still retained. Candidate identifiers are lowercased and checked against the eleven identifiers in the current payload. An unusable or illegal candidate triggers another Bedrock call, up to three total attempts. On retries, the runner appends the reminder: \emph{Your previous response could not be used. Return only one JSON object with a legal candidate\_id and action.} A run fails if all three attempts lack a legal candidate; it does not silently choose a default detector. The executor derives the authoritative action from the previous selection: \texttt{initialize} for the first choice, \texttt{keep} for an unchanged candidate, and \texttt{switch} otherwise. If the model reports a different action, the reported value and normalization flag remain in the decision artifact. \par \paragraph{Execution and audit trail.} The runner writes each controller's payload, normalized decision, raw model response, attempt traces, token usage, and timestamp into a \texttt{pretrain\_decisions} JSON file before launching any current-window trainer. Only the union of the controllers' selected candidates is trained; when a fixed TGN diagnostic was declared in advance, that candidate is added after selection and its scores are not returned to the agents. A shared choice is trained once and referenced by both trajectories. Training settings, graph adapters, and early stopping are set by the executor, not generated by the language model. The executor writes validation and hidden deployment metrics to evaluator-side artifacts, and later exposes a deployment outcome only when the delay rule makes it mature. Each trajectory stores the selected model, its training time and artifact path, and its observed outcome. This ordering permits an auditor to compare decision timestamps with trainer logs and to check that no current-window score entered the prompt. \par \paragraph{No-memory comparison.} In the no-memory condition, the runner keeps the same fixed instruction and eleven candidate cards but sets \texttt{memory\_mode} to \texttt{no\_memory}, \texttt{previous\_selection} to null, and \texttt{matured\_history} to an empty array at every window. Consequently, the required reported action is \texttt{initialize} at every invocation. This ablation removes cross-window controller state while leaving the detector architectures, per-window training code, and evaluator unchanged; detector-internal temporal memory is reset with each training episode in both conditions. \par  \end{document}